\documentclass{article}
\usepackage{ijcai19}
\usepackage{times}
\usepackage{soul}
\usepackage{url}
\usepackage[utf8]{inputenc}
\usepackage[small]{caption}
\usepackage{graphicx}
\usepackage{amsmath}
\usepackage{booktabs}
\usepackage{multirow}
\usepackage[draft]{hyperref}
\usepackage{amsfonts}

\title{HorNet: A Hierarchical Offshoot Recurrent Network for Improving Person Re-ID via Image Captioning}

\author{
Shiyang Yan$^{1,2}$\and
Jun Xu$^{1,2}$\and
Yuai Liu$^{1,2}$\And
Lin Xu$^{1,2}$\footnote{Contact Author}
\affiliations
$^1$Nanjing Institute of Advanced Artificial Intelligence  \\
$^2$Horizon Robotics\\
\emails
elyotyan@gmail.com,
\{jun.xu, yuai.liu, lin01.xu\}@horizon.ai
}

\begin{document}

\maketitle

\begin{abstract}
Person re-identification (re-ID)  aims to recognize a person-of-interest across different cameras with notable appearance variance. Existing research works  focused on the capability and robustness of visual representation. In this paper, instead, we propose a novel hierarchical offshoot recurrent network (HorNet) for improving person re-ID via image captioning. Image captions are semantically richer and more consistent than visual attributes, which could significantly alleviate the variance. We use the similarity preserving generative adversarial network (SPGAN) and an image captioner to fulfill domain transfer and language descriptions generation. Then the proposed HorNet can learn the visual and language representation from both the images and captions jointly, and thus enhance the performance of person re-ID. Extensive experiments are conducted on several benchmark datasets with or without image captions, i.e., CUHK03, Market-1501, and Duke-MTMC, demonstrating the superiority of the proposed method. Our method can generate and extract meaningful image captions while achieving  state-of-the-art performance.
\end{abstract}

\section{Introduction}
\begin{figure}[htb]
	\centering
	\includegraphics[height=6cm,width=8cm]{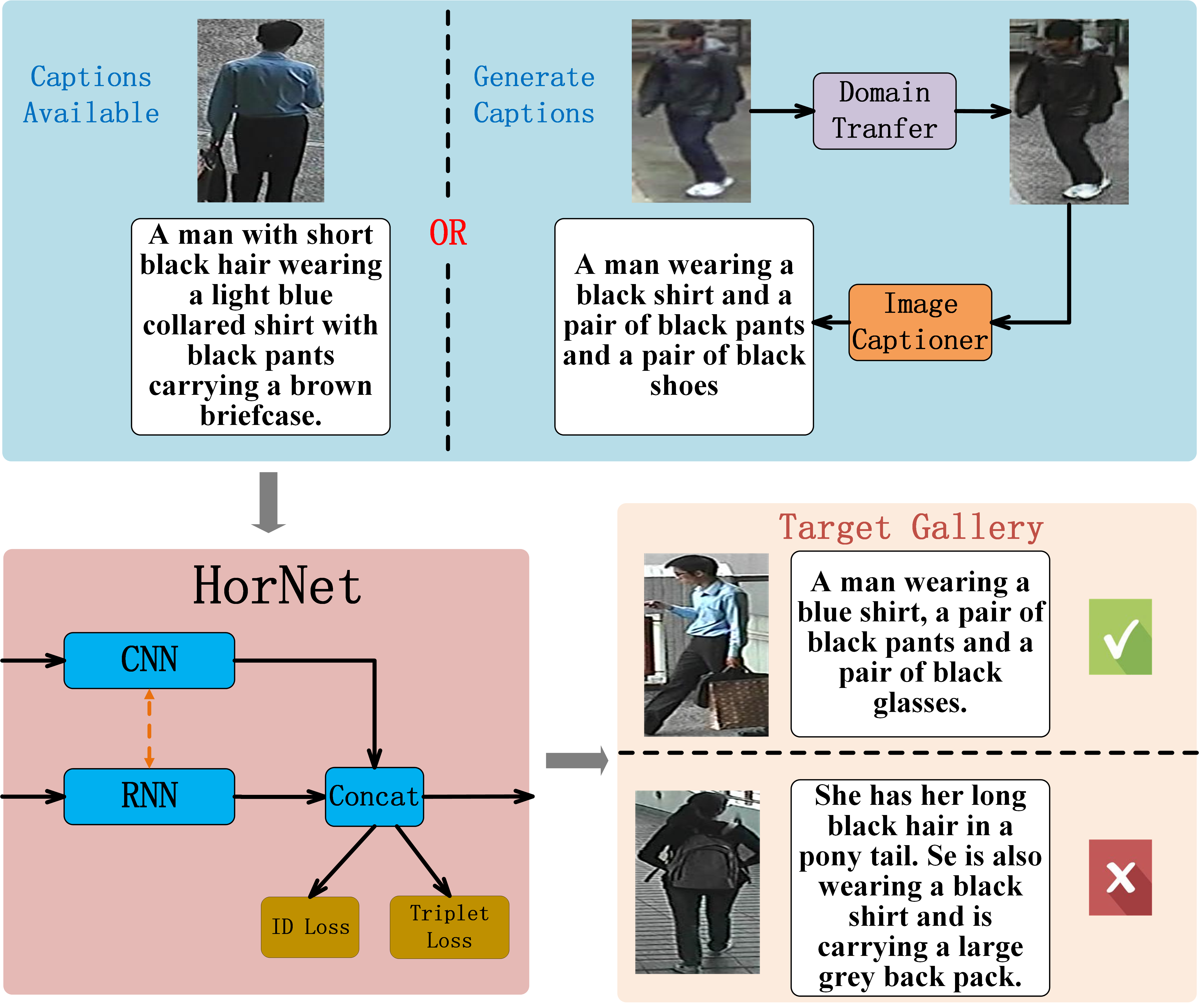}
		\vspace{-0.2cm}
	\caption{Schematic illustration of the proposed framework for person re-ID. Both images and captions are utilized for spotting a person-of-interest across different cameras.  For persons without captions, we first transfer all available images into a unified domain and then use image captioner to generate high-quality language description automatically. The HorNet simultaneously extracts visual representation from a given image and language description from the generated caption for the following person re-identification.}\label{system0}
	\vspace{-0.4cm}
\end{figure}

Person re-identification (re-ID) has become increasingly popular in the modern computer vision community due to its great significance in the research and applications of visual surveillance. It aims at recognizing a person-of-interest (query) across different cameras. The most challenging problem in re-ID is how to accurately match persons under intensive variance of appearances, such as human poses, camera viewpoints, and illumination conditions. Encouraged by the remarkable success in deep learning algorithms and the emergence of large-scale datasets, many advanced methods have been developed to relieve these vision-based difficulties and made significant improvements in the 
community \cite{li2017learning,su2017pose,chen2018improving}.

Recent years witness that the application of the various axillary information, such as human poses~\cite{su2017pose}, person attributes~\cite{schumann2017person} and language descriptions~\cite{chen2018improving}, can significantly boost the performance of person re-ID. These serve as the augmented feature representations for improving person re-ID. Notably, the image captions could provide a comprehensive and detailed footprint of a specific person. It is semantically richer than visual attributes. More importantly,  language descriptions of a particular person are often more consistent across different cameras (or views), which could alleviate the difficulty of the appearance variance in person re-ID task. 

Two significant barriers exist in applying the image captions for person re-ID. The first one is the increasing complexity to handle image captions. It is certain that language descriptions contain many redundant and fuzzy information, which could be a great challenge if not handled properly. Thus an effective learning approach for constructing a compact representation of language descriptions is of vital importance. Another one is the lack of description annotations for person re-ID task. Recently, \cite{li2017person} proposed the  CUHK-PEDES, which provides person images with annotated captions. The images from this dataset are collected from various person re-ID benchmark datasets such as  CUHK01~\cite{li2012human}, CUHK03~\cite{li2014deepreid}, Market-1501~\cite{zheng2015person},  and et al. However, the annotations are usually restricted to these datasets. In real-world applications, the person images normally do not have paired language descriptions. Thus, a method for automatically generating the high-quality semantic image captions to various real-world datasets is also urgently needed.

In this paper, we propose a novel hierarchical offshoot recurrent network (HorNet) for improving person re-ID via image captioning. 
Figure \ref{system0} illustrates the schematic illustration of our framework for person re-ID task. We first use the similarity preserving generative adversarial network (SPGAN)~\cite{deng2018image} to transfer the real-world images into a unified domain, which can significantly enhance the quality of the generated descriptions via the following image captioner~\cite{AnejaConvImgCap17}. Then both of the images and generated captions are used as the input to the HorNet. The HorNet has two sub-networks to handle the input images and captions, respectively. For images, we utilize mainstream CNNs (i.e., Resnet50) to extract the visual features. For captions, we develop a two-layer LSTMs module with a discrete binary gate in each time step. The gradient of the separate gates is estimated using $Gumbel\_sigmoid$~\cite{jang2016categorical}. This module dynamically controls the information flow from the lower layer to the upper layer via these gates. It selects the most relevant vocabularies (i.e., the correct or meaningful words), which are consistent with the input visual features. Consequently, HorNet can learn the visual representations from the given images and the language descriptions from the generated image captions jointly,  and thus significantly enhance the performance of person re-ID.
Finally, we verify the performance of our proposed method in two  scenarios, i.e., person re-ID datasets with and without image captions. Experimental results on several widely used benchmark datasets, i.e., CUHK03, Market-1501, and Duke-MTMC, demonstrate the superiority of the proposed method. Our method 
can simultaneously learn the visual and language representation from both the images and captions  while achieving a state-of-the-art recognition performance.

In a nutshell,  our main contributions in the present work can be summarized as  threefold:

  
(1)  We develop a  new  captioning module via image domain transfer and captioner in person re-ID system. It can generate high-quality language captions for given visual images. 

(2) We propose a novel hierarchical offshoot recurrent network (HorNet) based on the generated images captions, which learns the visual and language representation jointly. 

(3) We verify the superiority of our proposed method on person re-ID task. State-of-the-art empirical results are achieved on the three commonly used benchmark datasets.

%
%
\section{Related Work}
The early research works on person re-ID mainly focus on the visual feature extraction. For instance, \cite{yi2014deep} split a pedestrian image into three horizontal parts and train three-part CNNs to extract features. Then the similarity between two images is calculated based on the cosine distance metric of their features. 
\cite{chen2018improving} use triplet samples for training the network,  considering not only the samples of the same person but also the samples of different people.
\cite{liu2016multi} proposes a multi-scale triplet CNN for person re-ID. Due to recently released large-scale benchmark dataset, e.g., CUHK03~\cite{li2014deepreid}, Market-1501~\cite{zheng2015person}, many researchers try to learn a deep model based on the identity loss for person re-ID. \cite{zheng2016person} directly uses a conventional fine-tuning approach and outperforms many previous results. Also, recent research~\cite{zheng2017discriminatively} proves that a discriminative loss, combined with the verification loss objective, is superior.

Several recent research has endeavored to use auxiliary information to aid the feature representation for the person re-ID. Some research~\cite{su2017pose,zhao2017deeply} relies on the extra information of the person's poses for person re-ID. They leverage the human parts cues to alleviate the pose variations and learn robust feature representations from both the global and local image regions. Another type of auxiliary information, attributes of a person, has been used in person re-ID~\cite{lin2017improving}. However, these methods all rely on the attribute annotations, which are normally hard to collect in real-world applications. \cite{schumann2017person} uses automatically detected attributes and visual features for person re-ID. The attribute detector is trained on another dataset which contains the attribute annotations.

The relationship between visual representations and language descriptions has long been investigated. It has attracted high attention in tasks such as image captioning~\cite{yan2018image}, visual question answering. Associating person images and their corresponding language descriptions for the person searching has been proposed in~\cite{li2017person}. Several research works employ the language descriptions as complementary information, together with visual representations, for person re-ID. \cite{chen2018improving} exploit natural language descriptions as additional training supervision for effective visual features. \cite{yan2018person} propose to combine the language descriptions and image features and fuse them for the person re-ID task.
\begin{figure*}[htb]
	\centering
	\includegraphics[width=0.8\linewidth]{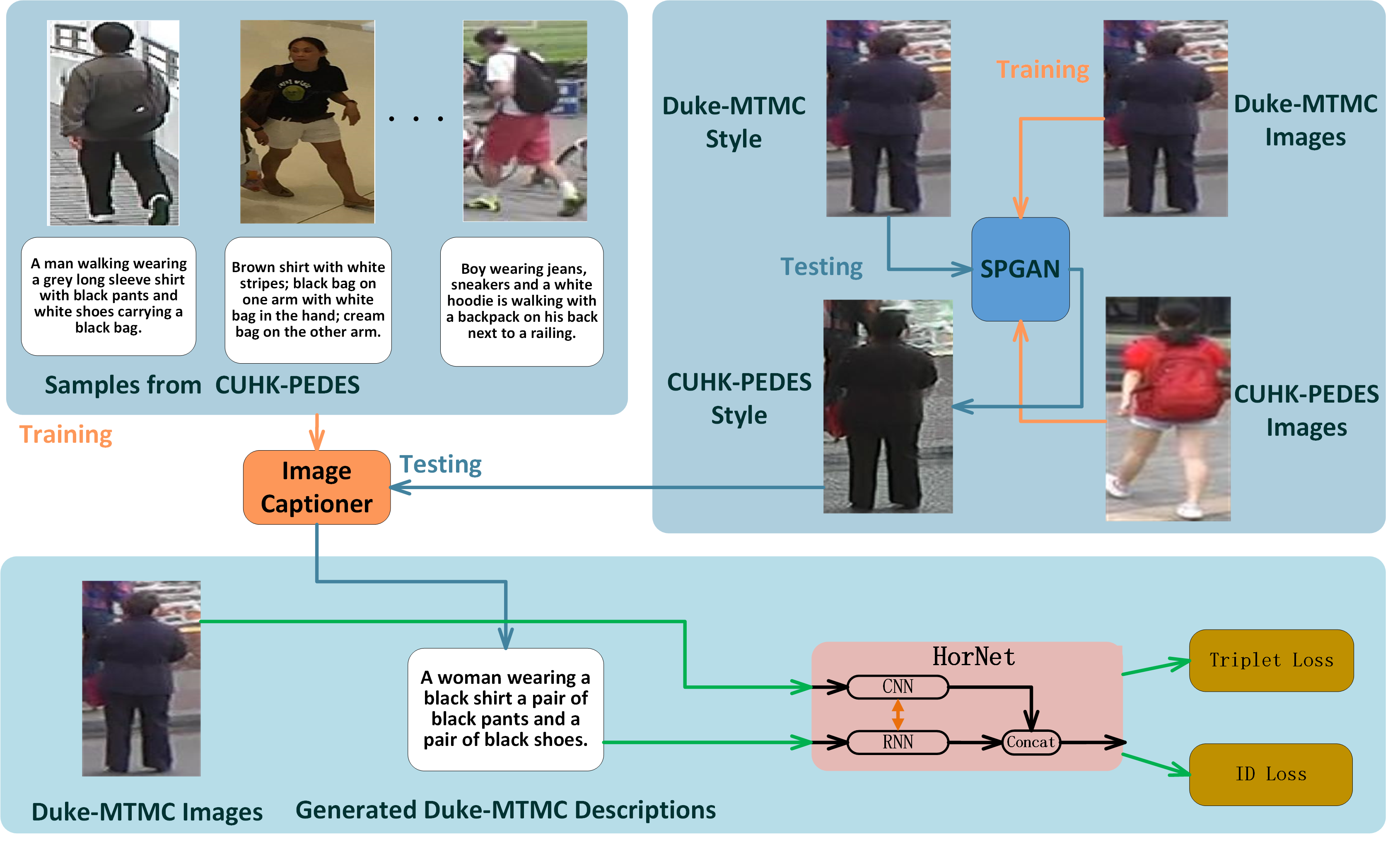}\\
	\caption{The pipeline of the proposed method for extending the language descriptions to the datasets without annotations: specifically, we first train an image captioning model on the CUHK-PEDES dataset. Then we transfer the image style of the Duke-MTMC dataset to the CUHK-PEDES style. The transferred the Duke-MTMC images with the CUHK-PEDES style are used to generate more precise language descriptions. Finally, we use the generated descriptions and the original Duke-MTMC images for person re-ID.}\label{system2}
	\vspace{-0.4cm}
\end{figure*}
Previous language models encode the sentences using either Recurrent Neural Networks (RNNs) language encoder or Convolutional Neural Networks (CNNs) encoder. Recent research~\cite{bahdanau2014neural} employ the attention mechanism for these language models by looking over the entire sentence and assigning weights to each word independently. Especially, RNNs with attention have been widely applied in machine translation, image captioning, speech recognition and Question and Answering (QA). The attention mechanism allows the model to look over the entire sequence and pick up the most relevant information. Most of the previous attention mechanism employs a similar approach to~\cite{bahdanau2014neural}, where the neural model assigns soft weights on the input tokens. Recently,~\cite{ke2018focused} proposes a Focused Hierarchical Encoder (FHE) for the Question Answering (QA), which consists of multi-layer LSTMs with the discrete binary gates between each layer. Our HorNet also utilizes the discrete gate but with a very different mechanism and purpose. We aim to eliminate redundant information or incorrect language tokens, while they tried to answer the  question.

\section{Our Method}
\subsection{Improving Person Re-ID via Image Captioning}
The image caption of a specific person is semantically rich and can provide complementary information for the visual representations. However,  the handcrafted descriptions of a person image are hard to collect due to the annotation difficulties in real-world person re-ID applications. We propose a method to generalize the language descriptions accurately from a dataset with image captions to others without such captions. The whole scheme of our approach is illustrated in Figure~\ref{system2}.  Given images with captions, i.e., the CUHK-PEDES dataset, we use SPGAN to transfer arbitrary image to the CUHK-PEDES style. The SPGAN is proposed to improve image-to-image domain adaptation by preserving both the self-similarity and domain-dissimilarity for person re-ID. 
We utilize it in our case as in \cite{deng2018image} to transfer the image domain (or style) of the un-annotated datasets. Then we train an image captioner \cite{AnejaConvImgCap17} to generate image descriptions automatically on the transferred datasets.
The visualization of the domain transfer process and corresponding generated captions are illustrated in Figure \ref{spgan}. It is clear that the transferred images have more accurate language descriptions. However, the generated sentences, which are based on the domain-translated images, still contain some incorrect keywords and redundant information. The proposed HorNet, which contains the discrete binary gates, can select the most relevant language tokens with the visual features, and thus provide a good solution for the issue.
\begin{figure*}[htb]
	\centering
	\includegraphics[height=7cm,width=15cm]{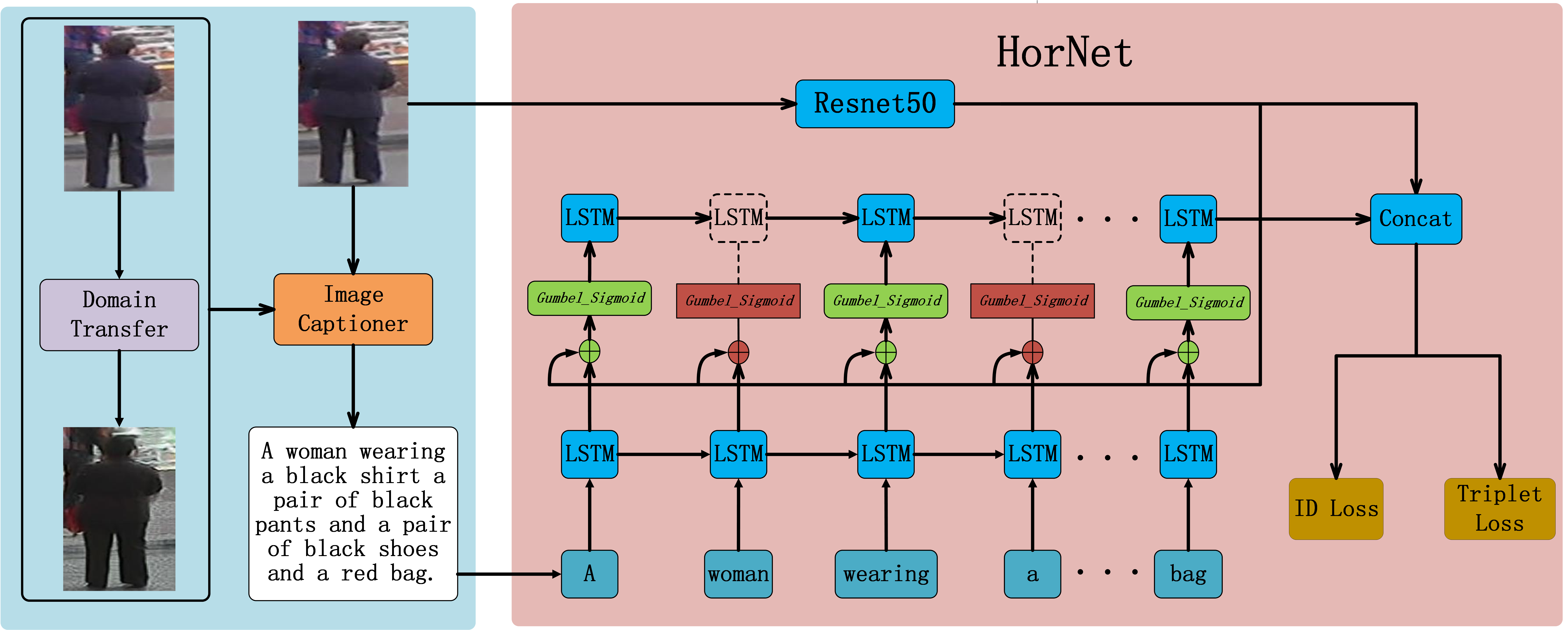}\\
	\caption{The pipeline of the proposed method. We first use the domain transfer technique (i.e., SPGAN) to transfer all available training images into a unified domain (or style).  This preprocess can significantly enhance the quality of the generated language descriptions via the following image captioner. 
		In the structure of the HorNet, the information flow from the lower-layer LSTM to the upper-layer LSTM is controlled by the discrete binary gates. The discrete binary gates are determined by the corresponding visual features and the hidden units from the lower LSTM layer. The gradient of the discrete gates is then estimated via the $Gumbel\_sigmoid$ function. The red circles in the figure indicate the closeness of the gates, while the yellow circles mean the gates are open. The final concatenate representation from both visual and language features are employed for person re-ID through ID loss and Triplet loss objectives simultaneously.}\label{system}
	\vspace{-0.4cm}
\end{figure*}
\subsection{The Proposed HorNet Model}
To facilitate the visual representations of a person in the person re-ID task,  we propose the HorNet to learn the visual features and the corresponding language description jointly.  The HorNet adds a branch to the CNNs (i.e., Resnet-50) with two-layer LSTMs and a discrete or continuous gate between each layer at every time step. The lower-layer LSTM handles the input languages while the upper-layer LSTM selects the relevant language features via the gates. Finally, the last hidden state of the upper-layer LSTM is concatenated with the visual features extracted via Resnet-50 to generate a compact representation. The objective function of the HorNet consists of two parts: the identification loss and the triplet loss, which are trained jointly to optimize the person re-ID model.

We present the pipeline of our proposed method in Figure~\ref{system}.  The input to the HorNet consists of two parts, i.e.,  the image and the corresponding language descriptions.  Let the language descriptions be processed by a two-layer LSTM model.  The bottom layer is a normal LSTM network, which reasons on the sequential input of the language descriptions. More formally, let the $D = (d_1, d_2, ..., d_n)$ be the input of description, $h_t$ be the hidden state, $c_t$ be the LSTM cell state at time $t$. The term $E = (e_1, e_2, ..., e_n)$, where $e_t = Word\_Embedding(d_t), t = 1, 2, ..., n$, denotes the word embedding  of the input description. In our research work, we use a linear embedding for the input language tokens. Hence, the bottom $LSTM$ layer can be expressed in   Equation (\ref{bottom1}).
\begin{equation}\label{bottom1}
h_t^{l}, c_t^{l} = LSTM(e_t, h_{t-1}^{l}, c_{t-1}^{l}), 
\end{equation}
where function $LSTM$ denotes the compact form of the forward pass of an LSTM unit with a forget gate as:
\begin{equation}\label{bottom}
\begin{aligned}
&  i_t^l = \sigma({W_{xi}^l}\ast{e_t}+{U_{hi}^l}\ast{h_{t-1}^l}+b_i^l), \\
&  f_t^l = \sigma({W_{xf}^l}\ast{e_t}+{U_{hf}^l}\ast{h_{t-1}^l}+b_f^l), \\
&  o_t^l = \sigma({W_{xo}^l}\ast{e_t}+{U_{ho}^l}\ast{h_{t-1}^l}+b_o^l), \\
&  g_t^l = \sigma({W_{xc}^l}\ast{e_t}+{U_{hc}^l}\ast{h_{t-1}^l}+b_c^l), \\
&  c_t^l = f_t \cdot c_{t-1}^l + i_t^l \cdot g_t^l ,\\
&  h_t^l = o_t \cdot \phi(c_t),
\end{aligned}
\end{equation}
where $e_t \in \mathbb{R}^{d}$ denotes input vector, 
$f_t^l \in \mathbb{R}^{h}$ is forget gate's activation, 
$i_t^l \in \mathbb{R}^{h}$is input gate's activation, 
$o_t^l \in \mathbb{R}^{h}$ is  output gate's activation, 
$c_t^l \in \mathbb{R}^{h}$ is cell state vector, and 
$h_t^l \in \mathbb{R}^{h}$ is hidden state of the LSTM unit $l$.
$W \in \mathbb{R}^{h \times d},  U \in \mathbb{R}^{h \times h}$ and $b \in \mathbb{R}^{h}$ are  weight matrices and bias vector parameters which need to be learned during training. 
The activation $\sigma$ is sigmoid function and the operator $\ast$ denotes the Hadamard product (i.e., element-wise product).

The boundary gate controls the information from the lower layer to the upper layer. The boundary gate $z_t$ is estimated with $Gumbel\_sigmoid$, which is derived directly from the $Gumbel\_softmax$ proposed in \cite{jang2016categorical}.

The $Gumbel\_softmax$ replaces the $argmax$ in the $Gumbel$-Max Trick with the following Softmax function:
\begin{equation*}\label{gumbelsoftmax}
Gumbel\_softmax(\pi_i) = \frac{exp(log(\pi_i + g_i)/\tau)}{\sum_{j=1}^{K}exp(log(\pi_j + g_j)/\tau)},
\end{equation*}
where $g_1, ..., g_k$ are $i.i.d.$ sampled from the distribution $Gumbel (0,1)$, and $\tau$ is the temperature parameter. $K$ indicates the dimension of the generated Softmax vector (i.e., the number of categories).

To derive the $Gumbel\_sigmoid$, we firstly re-write the Sigmoid function as a Softmax of two values: $\pi_i$ and 0, as in  the following Equation (\ref{sigmoid}).
\begin{equation}\label{sigmoid}
\begin{aligned}
sigm(\pi_i)& =\frac{1}{(1 + exp({ - \pi_i}))} =\frac{1}{(1 + exp({0 - \pi_i}))}   \\
& =  \frac{1}{1 + {exp(0)}/{exp(\pi_i)}} \\
& =\frac{exp({\pi_i})}{(exp({\pi_i}) + exp(0))}.
\end{aligned}
\end{equation}

Hence, the $Gumbel\_sigmoid$ can be written as in the following Equation (\ref{gumbelsigmoid}).
\begin{equation}\label{gumbelsigmoid}
\begin{aligned}
& Gumbel\_sigmoid(\pi_i) =  \\
&\frac{exp(log(\pi_i+g_i)/{\tau})} {exp({log(\pi_i+g_i)/{\tau}})+exp({log(g^\prime)/{\tau}})},
\end{aligned}
\end{equation}
where $g_i$ and $g^\prime$ are independently sampled from the distribution $Gumbel (0,1)$.

Thus, the upper-layer LSTM inputs are the gated hidden units of the lower-layer, which can be expressed as the following Equation (\ref{upper2}) and Equation (\ref{upper1}). In our experiments, all the soft gates $z_t$ are estimated using the $Gumbel\_sigmoid$ with a constant $\tau$ of $0.3$. 
\begin{equation}\label{upper2}
z_t = Gumbel\_sigmoid(Concat(h_t^{l}, F)),
\end{equation}
\begin{equation}\label{upper1}
  h_t^{l+1}, c_t^{l+1} = LSTM(h_t^l * z_t, h_{t-1}^{l+1}, c_{t-1}^{l+1}),
\end{equation}
where $F$ denotes the deep visual features of the images extracted via CNNs (i.e., Resnet-50) and the $Concat$ indicates the features concatenation operation. 

To obtain a discrete value (i.e., language tokens selection), we also set the hard gates $z_t = \widetilde{y_i}$ in Equation (\ref{upper2}).
\begin{equation}\label{gumbelsigmoid2}
\widetilde{y_i} = \begin{cases}
1 \qquad   y_i >= 0.5,  \\
0 \qquad   otherwise.
\end{cases}
\end{equation}

Finally,  we forward the last hidden unit of the language branch  to form a compact representation by using a concatenation operation with the corresponding visual features $F$ as, 
\begin{equation}
{f}  = Concat(h_n^{l+1}, F).
\end{equation}  
Our loss function is the combination of the identification loss and triplet loss objectives, which can be expressed in the following Equation  (\ref{finalloss}) as, 
\begin{equation}\label{finalloss}
\mathcal L =  \mathcal L_{ID}+ \mathcal L_{Triplet}, 
\end{equation}
where $\mathcal L_{ID}$ is a $K$-class cross entropy loss parametered by $\theta$. It treats each ID number of person as an individual class as, 
\begin{equation}
\mathcal L_{ID} =  - \sum\nolimits_k^K {{y_k}} \log (\frac{{{e^{{\theta _k}f}}}}{{\sum\nolimits_j^K {{e^{{\theta _j}f}}} }}), 
\end{equation}
and $\mathcal L_{Triplet}$ denotes the Triplet loss  as, 
\begin{equation}\label{loss}
\begin{aligned}
\mathcal {L}_{Triplet}=  &  {max} ({\| f \left(x_a\right)- f \left(x_p\right)\|}^{2}- \\
 & {\| f \left(x_a\right)- f \left(x_n\right)\|}^{2} 
  +\alpha ,0 ), 
\end{aligned}
\end{equation}
where $x_a$ is the anchor, $x_p$ indicates the positive example, and $x_n$ is the negative sample. The $\alpha$  means margin.


\section{Experiments}
\begin{table*}[htb]
\centering
\resizebox{0.85\linewidth}{!}{
\begin{tabular}{|c||c|c|c|c||c|c|c|c||c|c|c|c|c|}
\hline
\multirow{2}{*}{Methods} & \multicolumn{4}{|c||}{Market-1501} & \multicolumn{4}{|c||}{CUHK03 Detected} & \multicolumn{4}{|c|}{CUHK03 Labeled}  \\
\cline{2-13}
& mAP & top-1 & top-5 & top-10 & top-1 & top-5 & top-10 & top-20 & top-1 & top-5 & top-10 & top-20  \\
\hline
Identification Loss & 65.5 & 82.4 & 92.9 & 95.3 &  72.4 & 89.0 & 93.0 & 96.0 &  79.3 & 90.6 & 92.3 & 93.0  \\
Identification + Triplet Loss &  71.4  & 86.3  & 95.1   &  \textbf{96.9}   &  88.3 & 98.0 & 98.9 & 99.3 & 92.2 & 99.2 & 99.6 & 99.8 \\
Identification Loss + HorNet & 65.3  &  82.6 & 93.0 &  95.7  & 80.0 & 91.0 & 92.4 & 93.1 & 81.9 & 92.2 & 93.1 & 93.6  \\
Identification + Triplet Loss + HorNet & 73.3   & 88.6  &  \textbf{95.2}   &  96.7  &  91.5 & 98.5 & \textbf{99.2} & \textbf{99.4}  & 92.4 & 98.9 & 99.5 & 99.7\\
HorNet  + Rerank & \textbf{85.6} & \textbf{91.0} & 94.7 & 95.9 & \textbf{95.0} &  \textbf{98.8} & 99.0 & \textbf{99.4} & \textbf{97.1}  & \textbf{99.4}  &  \textbf{99.7} & \textbf{99.8} \\
\hline
\end{tabular}
}
	\vspace{-0.2cm}
\caption{Ablation study results on Market-1501, CUHK03 Detected, and CUHK03 Labeled datasets.}\label{table:compare}
	\vspace{-0.4cm}
\end{table*}
\begin{table}[htb]
	\centering
	\scalebox{.7}[.8]{
		\begin{tabular}{|c|c|c|c|c|c|}
			\hline
			\multirow{2}{*}{Methods}  & \multicolumn{2}{|c|}{CUHK03 (Detected)} &  \multicolumn{2}{|c|}{CUHK03 (Labeled)} \\
			\cline{2-5}
			& top-1 & top-5  & top-1 & top-5  \\
			\hline
			MSCAN~\cite{li2017learning} & 68.0 & 91.2 &  74.2 & 94.3  \\
			SSM~\cite{bai2017scalable} & 72.7 & 92.4 &  76.6 & 94.6  \\
			k-rank~\cite{zheng2017unlabeled} & 58.5 & -  & 61.6 & - \\
			JLMT~\cite{li2017person} & 89.4 & 98.2 &  91.5 & 99.0 \\
			Deep Person~\cite{bai2017deep} & 89.4 & 98.2 &  91.5 & 99.0  \\
			SVDNet~\cite{sun2017svdnet} & 81.8 & 95.2  & -& - \\
			MuDeep~\cite{fu2017multi} & 75.6 & 94.4 & - &  76.9 \\
			\hline
			DSE~\cite{chang2018joint} & 66.8 & 92.9 & - & - \\
			ACRN~\cite{schumann2017person} & 62.6 & 89.7 &  - & -  \\
			VL~\cite{yan2018person} & - & - & 81.8 & 98.1 \\
			ILA~\cite{chen2018improving} & 90.9 & 98.2  & 92.5 & 98.8  \\
			\hline
			\textbf{HorNet + Rerank} & \textbf{95.0} &  \textbf{98.8}  & \textbf{97.1}  & \textbf{99.4}  \\
			\hline
		\end{tabular}
	}
\vspace{-0.2cm}
\caption{Comparison with baselines on the CUHK03 dataset.}\label{table:cuhkclassic}
	\vspace{-0.1cm}
\end{table}
\begin{table}[htb]
	\centering
	\scalebox{.8}[.8]{
		\begin{tabular}{|c|c|c|c|c|c|}
			\hline
			\multirow{2}{*}{Methods} & \multicolumn{4}{|c|}{Market-1501}  \\
			\cline{2-5}
			& mAP & top-1 & top-5 & top-10  \\
			\hline
			MSCAN~\cite{li2017learning} & 57.5 & 80.3 & - & \\
			SSM~\cite{bai2017scalable} & 68.8 & 82.2 & - & - \\
			k-rank~\cite{zheng2017unlabeled} & 63.4 & 77.1 & - & - \\
			SVDNet~\cite{sun2017svdnet} & 62.1 & 82.3 & - & - \\
			DPLAR~\cite{zhao2017deeply} & 63.4 & 81.0 &- & - \\
			PDC~\cite{su2017pose} & 63.4 & 84.1 & - & - \\
			JLMT~\cite{li2017person} & 65.5 & 85.1 & - & - \\
			D-person~\cite{bai2017deep} & 79.6 & 92.3 & - & - \\
			TGP~\cite{almazan2018re} & 81.2 & 92.2 & - & - \\
			\hline
			DSE~\cite{chang2018joint} & 64.8 & 84.7 & -&- \\
			ACRN~\cite{schumann2017person} & 62.6 & 83.6 & - & - \\
			ILA~\cite{chen2018improving} & 81.8 & \textbf{93.3} & - & - \\
			\hline
			\textbf{HorNet + Rerank} & \textbf{85.8} & 91.0 & \textbf{94.2} & \textbf{97.4} \\
			\hline
		\end{tabular}
	}
\vspace{-0.2cm}
\caption{Comparison with baselines on the Market-1501 dataset.}\label{table:market}
	\vspace{-0.4cm}
\end{table}

\begin{table}[htb]
	\centering
	\scalebox{.65}[.8]{
		\begin{tabular}{|c|c|c|c|c|c|}
			\hline
			\multirow{2}{*}{Methods}  & \multicolumn{4}{|c|}{Duke-MTMC} \\
			\cline{2-5}
			& mAP & top-1 & top-5 & top-10  \\
			\hline
			BoW + Kissme~\cite{zheng2015scalable} & 12.2 & 25.1 & - & -\\
			LOMO + XQDA~\cite{liao2015person} & 17.0 & 30.8 &- & -\\
			Verification + Identification~\cite{zheng2017discriminatively} & 49.3 & 68.9 &- &- \\
			PAN~\cite{zheng2018pedestrian} & 51.5 & 71.6 & -&- \\
			PAN + Rerank~\cite{zheng2018pedestrian} & 66.7 & 75.9 &- &- \\
			FMN~\cite{ding2017let} & 56.9 & 74.5 &- &- \\
			FMN + Rerank~\cite{ding2017let} & 72.8 & 79.5 &- &- \\
			D-person~\cite{bai2017deep} & 64.8 & 80.9 & -&- \\
			SVDNet~\cite{sun2017svdnet} & 56.8 & 76.7 & - & - \\
			\hline
			APR~\cite{lin2017improving} & 51.9 & 71.0 & -&- \\
			ACRN~\cite{schumann2017person} & 52.0 & 72.6 & 88.9 &  91.5\\
			\hline
			Resnet50 + BERT + Rerank~\cite{Jacob2018bert}& 78.8 & 84.1 & 90.0 &  92.2\\
			\hline
			Identification Loss & 54.6 & 72.5 & 84.4 & 88.7 \\
			Identification Loss + HorNet (Without Domain Transfer) & 52.5 & 71.1 & 82.6 & 87.7 \\
			Identification Loss + HorNet (With Domain Transfer) & 58.4 & 74.3 & 87.3 & 90.8 \\
			HorNet (With Domain Transfer) & 60.4 & 76.4 & 88.1 & 90.5 \\
			\textbf{HorNet + Rerank} & \textbf{79.2} & \textbf{84.4} & \textbf{90.2} & \textbf{92.5} \\
			\hline
		\end{tabular}
	}
\vspace{-0.2cm}
	\caption{Comparison with  baselines on the Duke-MTMC dataset.}\label{table:duke}
	\vspace{-0.4cm}
\end{table}
We evaluated the proposed methods on  person re-ID datasets such as CUHK03~\cite{li2014deepreid}, Market-1501~\cite{zheng2015person} and Duke-MTMC~\cite{ristani2016performance}. There are two types of experiments: with and without description annotations. For CUHK03 and Market-1501, the description annotations can be directly retrieved from  the CUHK-PEDES dataset~\cite{li2017person}. Hence, we evaluated the proposed method for person re-ID by using these annotations.  However, Duke-MTMC lacks the language annotations. We used an image captioner to generate language descriptions, which are used to jointly optimize the proposed HorNet.

\subsection{Implementation Details}

For the HorNet, the embedding dimension of word vectors is $512$. The dimension of the hidden unit of the LSTM is also $512$. Since we used a fully connected layer to process the last hidden unit of the upper-layer LSTM and consequently reduced its dimension to 256,  the dimension of the final representation for vision and language is $2304$ (i.e., $256 + 2048$), in which the dimension of the visual features is $2048$.

\subsection{Experiments with CUHK-PEDES Annotations}
\begin{figure}[htb]
  \centering
  \includegraphics[width=\linewidth]{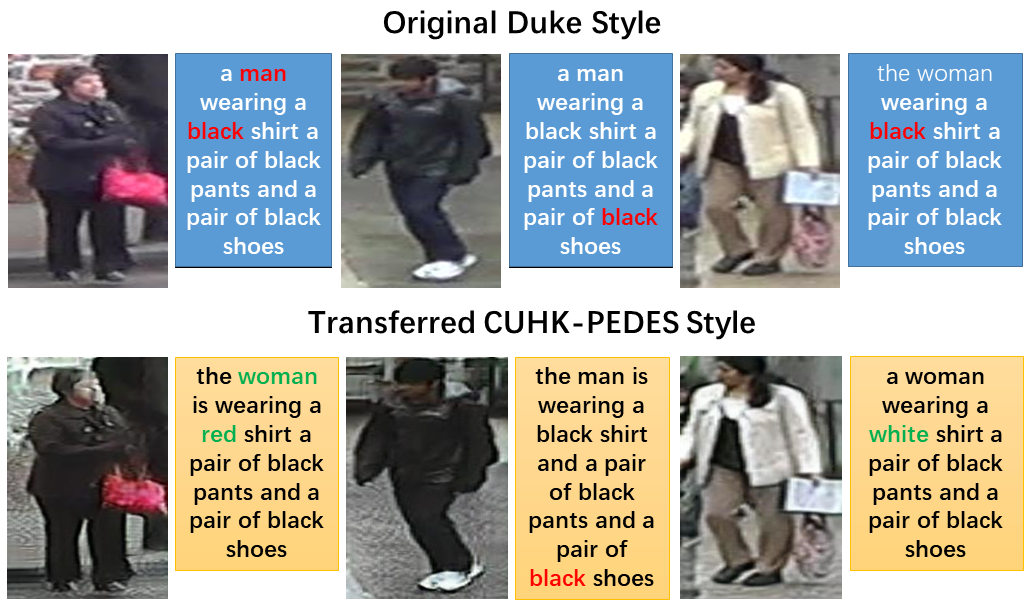}\\
  \caption{The visualization of the domain transfer process and corresponding generated captions. The red keywords indicate incorrect descriptions while the green words mean the correct keywords. }\label{spgan}
  \vspace{-0.4cm}
\end{figure}
To first verify the effect of the language descriptions in person re-ID, we augmented two standard person re-ID datasets, which are CUHK03 and Market-1501, by using annotations from CUHK-PEDES dataset. The annotations of  CUHK-PEDES are initially developed for cross-modal text-based person search. Since the persons in  Market-1501 and CUHK03 have many similar samples and only four images of each person in these two datasets have language descriptions, we annotated the unannotated images in those datasets with the language information from the same ID, which is the same as the protocol in \cite{chen2018improving}.

We first evaluated the proposed method on CUHK03 dataset, using the classical evaluation protocols~\cite{li2014deepreid}.  We tested the baseline method which uses the identification loss based on the Resnet-50 model, with $72.4$ CMC top-1 accuracy in the CUHK03 detected images. The CMC top-1 result is raised to $80.0$ by augmenting the language descriptions. A similar phenomenon can be seen in the CUHK03 labeled images. We performed an ablation study to verify the effect of the different part in HorNet. The experimental results are presented in Table~\ref{table:compare}. The proposed HorNet with reranking technique can achieve the best performance on Market-1501, CUHK03 detected images,  and CUHK03 labeled images datasets. 


The comparison with other state-of-the-art methods are listed in Table~\ref{table:cuhkclassic}. Specifically, we compared the proposed HorNet with other methods which employ auxiliary information, which include Deep Semantic Embedding (DSE)~\cite{chang2018joint}, ACRN~\cite{schumann2017person}, ACRN~\cite{schumann2017person}, Vision and Language (VL)~\cite{yan2018person}, Image-language Association (ILA)~\cite{chen2018improving}. ACRN applies axillary attribute information to aid the person re-ID. DSE, ACRN, VL and ILA all employ the external language descriptions for person re-ID. Among them, VL uses a vanilla LSTM or CNN language encoding model, which is discriminatively poorer than our HorNet, since the proposed HorNet uses discrete gates to select useful information for person re-ID. ILA uses the same training and testing protocol but with a more complex model. Our model can also be combined with
 various metric learning techniques, including the Rerank proposed in~\cite{zhong2017re}. We also employed the Rerank to post-process our features, with improved results. Overall, our HorNet performs much better than the ILA on the CUHK03 classic evaluation protocol, achieved $97.1\%$ CMC top-1 accuracy, with a $4.6\%$ raise over the ILA. We also conducted experiments on the Market-1501 dataset, the results are presented in Table~\ref{table:compare} and Table~\ref{table:market}. A similar phenomenon to those of CUHK03 can be seen in Table~\ref{table:market}, with $85.8\%$ mAP result.

\subsection{Experiments on the Duke-MTMC Dataset (Without Captions)}
In a realistic person re-ID system, language annotations are rare and hard to get. Hence, we want to see if the automatically generated language descriptions can boost the performance of a person re-ID system. We chose a more challenging and realistic dataset, i.e., Duke-MTMC~\cite{ristani2016performance} to verify this assumption. Firstly, we trained an image captioning model based on the CUHK-PEDES dataset by using the convolutional image captioning model, which has released code and good performance~\cite{AnejaConvImgCap17}. We split
 the CUHK-PEDES images into two splits: 95\% for training and 5\% for validation. We used the early stopping technique to train the image captioning model and achieved 35.4 BLEU-1, 22.4 BLEU-2, 15.0 BLEU-3, 9.9 BLEU-4, 22.3 METEOR, 34.2 ROUGE\_L and 22.1 CIDEr results on the validation set. Subsequently,
  we used the trained image captioning model to generate language descriptions for the Duke-MTMC dataset. However, we found that the generated descriptions are not discriminative enough, as shown in Figure~\ref{spgan}. There are many incorrent or imprecise  keywords in the language descriptions. Also, we tested the performance by augmenting the Duke-MTMC with the generated descriptions and the results turned out to be poor, even worse than the baselines, only with $52.5 \%$ mAP result, as shown in Table \ref{table:duke}.
  The cause of this phenomenon is the poor generalization capability of the image captioning model, especially when there is a domain difference between two diverse datasets. To alleviate this problem, we used the SPGAN~\cite{deng2018image} to transfer the image style of the Duke-MTMC to the CUHK-PEDES. The generated language descriptions are with much better quality, as presented in Figure~\ref{spgan}. The results from the augmentation with the generated language descriptions on the transferred Duke-MTMC images are much better than that provided by the simple visual features, with $60.4\%$ mAP result on Duke-MTMC. To prove the superiority of the HorNet, we also use BERT~\cite{Jacob2018bert} to replace HorNet, but with a poorer performance. Furthermore, we also implement a Rerank~\cite{zhong2017re} to boost the final recognition performance and achieved $79.2\%$ mAP result. 

\section{Conclusions}
In this paper,  we developed a language captioning module via image domain transfer and captioner techniques in person re-ID system. It can generate high-quality language descriptions for visual images, which can significantly compensate for the visual variance in person re-ID. 
Then we proposed a novel hierarchical offshoot recurrent network (HorNet) for improving person re-ID via such an automatical image captioning module.  It can learn the visual and language representation from both images and the generated captions, and thus enhance the performance.  The experiments demonstrate promising results of our model on CUHK03, Market-1501 and Duke-MTMC datasets.  Future research includes a more robust language captioning module and advanced metric learning methods.

\bibliographystyle{named}
\bibliography{835}

\end{document}